\documentclass[letterpaper, 10 pt, conference]{ieeeconf}  

\IEEEoverridecommandlockouts                              

\usepackage{graphicx}
\usepackage{float}
\usepackage{xcolor}
\usepackage{url}
\usepackage{hyperref}

\usepackage{listings} 

\title{\LARGE \bf
Bi-Manual Manipulation and Attachment via Sim-to-Real Reinforcement Learning
}

\author{Satoshi Kataoka$^{1}$, {\small Seyed} Kamyar {\small Seyed} Ghasemipour$^{1}$, Daniel Freeman$^{1}$, and Igor Mordatch$^{1}$%
\thanks{$^{1}$Google Research}%
}

\begin{document}

\maketitle
\thispagestyle{empty}
\pagestyle{empty}

\begin{abstract}

Most successes in robotic manipulation have been restricted to single-arm robots, which limits the range of solvable tasks to pick-and-place, insertion, and objects rearrangement. In contrast, dual and multi arm robot platforms unlock rich diversity of problems that can be tackled, such as laundry folding and executing cooking skills.
However, developing controllers for multi-arm robots is complexified by a number of unique challenges, such as the need for coordinated bimanual behaviors, and collision avoidance amongst robots. Given these challenges, in this work we study how to solve bi-manual tasks using reinforcement learning (RL) trained in simulation, such that the resulting policies can be executed on real robotic platforms. Our RL approach results in significant simplifications due to using real-time (4Hz) joint-space control and directly passing unfiltered observations to neural networks policies. We also extensively discuss modifications to our simulated environment which lead to effective training of RL policies.

In addition to designing control algorithms, a key challenge is how to design fair evaluation tasks for bi-manual robots that stress bimanual coordination, while removing orthogonal complicating factors such as high-level perception. In this work, we design a ``Connect Task", where the aim is for two robot arms to pick up and attach two blocks with magnetic connection points. We validate our approach with two xArm6 robots and 3D printed blocks with magnetic attachments, and find that our system has 
100\% success rate at picking up blocks, and
65\% success rate at the ``Connect Task".
Our accompanying project webpage can be found at:
\href{https://sites.google.com/view/bimanual-attachment}{sites.google.com/view/bimanual-attachment}
\end{abstract}

\section{INTRODUCTION}
    
    
    
    Robot arms are currently capable of accomplishing interesting and useful manipulation tasks, and have made their way into many industries. However, most such successes have been limited to single-arm robots, which limit solvable tasks to pick-and-place and object rearrangement problems.
    In comparison, dual and multi arm robot platforms unlock a richer set of problems that can be tackled, such as insertion \cite{levine2016end} and laundry folding \cite{shepard2010cloth}, and have the potential to solve many more tasks. Using multiple arms can also simplify existing robot tasks, such as relaxing the need for extremely dexterous in-hand manipulation when two hands could be used.
    
    However, the control problem for multi-arm robots is challenging due to a number of unique difficulties. Methods that execute pre-planned trajectories are more likely to see drift from the intended plan in bi-arm systems over single-arm systems. This is due to noise, asynchronous control delays, and coordination errors being amplified by having two arms. Pre-designing controllers for bi-manual systems in end-effector or task space control is also challenging as one needs to consider complex inter-arm collision avoidance and coordination behaviors. For this reason, we use reinforcement learning (RL) trained in simulation as our method to solve bi-manual tasks. Our RL approach offers the following simplifications: 1) We directly use real-time (4Hz) joint space control (rather than end-effector control), which relieves the need for run-time collision libraries or planners, 2) We do not use explicit filtering, state estimation, or observation engineering. Instead we simply use raw robot observations that are fed to a single neural network policy.
    
    Aside from designing control algorithms, a significant challenge is to design fair evaluation tasks for bi-manual robots. Some tasks are either too easy, as they consist of independent sub-tasks that do not require coordination and could be done with one arm. And other existing tasks are difficult due to challenges that are orthogonal to bimanual control. As examples, towel-folding stresses perception and modeling cloth dynamics, while insertion stresses high precision. In this work, we offer a task that connects two (and potentially more) blocks together. The blocks are connected via magnets, which require less precision than other mechanisms such as pegs and holes. Using solid blocks, perception and dynamics modeling problems are likewise simplified. The difficulty of the task is instead focused on coordinating two arms and cannot be performed by one arm alone. This task is also a useful end in itself because it can become a foundational operation for more complex assembly tasks in the future.
    
    To test our ideas, we implement a bi-manual platform with two xArm6 robots and 3D printed blocks with magnets attached. We find that our system has 100\% success rate with the task of picking up blocks and holding them at target locations, and 50-65\% success rate of connecting two blocks together.

    The contributions of our work are two-fold: We offer a sim-to-real reinforcement learning strategy designed for real-world bi-manual robot control in mind, as well as a magnetic block connection task that specifically targets challenges of bi-arm coordination. We hope these two contributions encourage increased study of rich tasks afforded by bimanual robotic systems.
    

\section{Task Definitions}
    \label{sec:tasks}
    In this work, our goal is to study how to obtain controllers for bimanual object manipulation through reinforcement learning (RL), such that they can be executed on real-world robotic systems. In this section we describe two tasks designed to study our questions of interest.


    \subsection{Pickup Task}
    \label{sec:pickup_task}
    Our first task is a pickup task, a precursor to studying bimanual control. In the pickup task, each trial begins with a block that is randomly placed on the ground in the robot workspace, and the robot arm must pick up the block and move it to a specific position in the air.
    Each trial is 25 seconds long. In simulation, we set 1mm distance between the center of block and the target position as the success condition, and in the real world we relax this success threshold to 5mm distance due to perception noises.
    
    \subsection{Connect Task}
\begin{figure}[t]
        \centering
        \includegraphics[scale=.56]{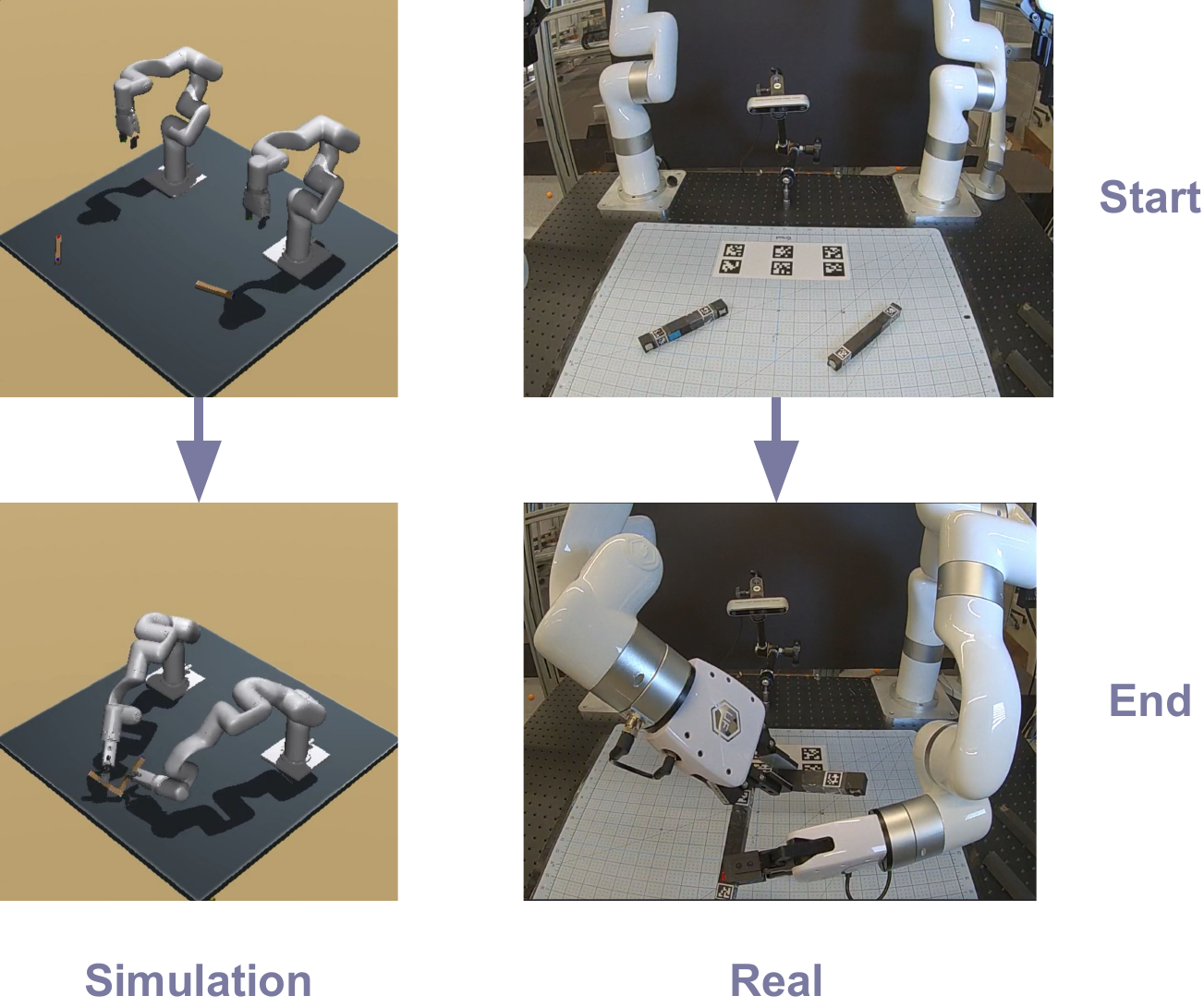}
        \caption{
            Visualization of simulated and real-world block \textit{Connect} Task.
        }
        \label{fig:connect_task}
    \end{figure}

    In the connect task, two blocks with magnetic connection points are placed on the ground, and two robotic arms must pick up the blocks and magnetically attach them. We designed this task as a minimal configuration that stresses bi-arm coordination and object manipulation. Despite its minimalism, increasing the number of magnetic blocks can support the creation of arbitrarily complex composed structures, which can lead to many intriguing avenues for future research. In our ``Connect Task", each trial is 25sec long. The success determination is different between simulation and real world. In simulation, we set 1mm distance and 0.05 radian orientations difference between the two magnets as the condition for success. In the real world, we set 1mm, 5mm, and 10mm as different levels of success and also set visual check for relative orientation as a success requirement,
    with a human operator checking the distance between the two magnets and the relative orientation between two blocks at the end of each trial. If the robot arms run into unsafe behavior in the real world, the human operator terminates the trial and marks it as a failure. Figure \ref{fig:connect_task} demonstrates the initial and goal states for the ``Connect Task" in the simulation and real domains.
 
\section{Physical Platform}
    We now describe the real-world setup where we solve the tasks described in Section \ref{sec:tasks}.
    We construct a physical work cell which consists of two robotics arms (UFACTORY xArm6 \cite{xarm6}), three cameras (Intel realsense D455 \cite{d455}) and two cuboid blocks with magnetic connection points. To estimate the position and orientation of the blocks, we use the AprilTag \cite{apriltag} tracking library. All blocks have 8 AprilTag markers attached to their surfaces. We also put 6 AprilTag markers on the workcell floor to track relative positions between the base and the blocks.
    We provide implementation details of these physical setups in the following subsections.
    
    \subsection{Robot Placement}
\begin{figure}[t]
        \centering
        \includegraphics[scale=.3]{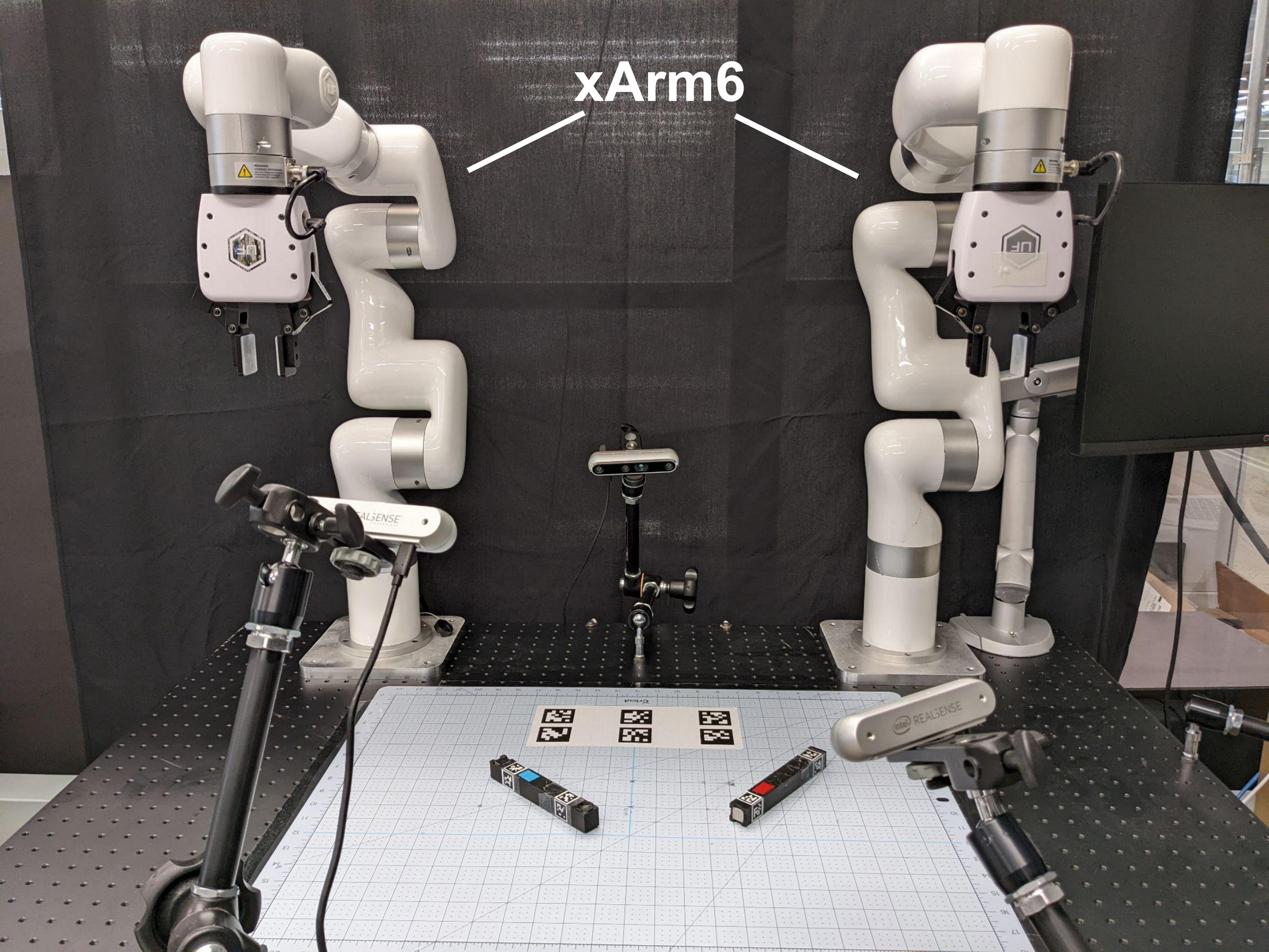}
        \caption{
                Dual xArm6 configuration.
        }
        \label{fig:robot_setup}
    \end{figure}

    In the workcell, two xArms are placed in the back of the workspace on the breadboard plate. To reduce the damage to the workcell tabletop, we put a 457.2mm x 609.6mm cutting mat at the center of the workcell. We place 6 Apriltag markers (37.5mm x 37.5mm) on the surface of the cutting mat.  The distance between the xArms is 711.2mm. The robots, breadboard, and base cutting mat can be seen in Figure \ref{fig:robot_setup}.
    
    \subsection{Block Specifications}
    In this work we use two cuboid blocks, one of each of the following types.
        \paragraph{} The three dimensional size of the first block is 19mm x 19mm x 152mm. Two magnets (12mm x 12mm) of opposite polarization are embedded into one of the long sides (19mm x 152mm) of the block.
        \paragraph{} The three dimensional size of the second block is also 19mm x 19mm x 152mm, but with a difference magnet arrangement. Two magnets of opposite polarization are embedded on the block, one on each of the two square sides (19mm x 19mm).
    
\begin{figure}[t]
        \begin{minipage}{0.22\textwidth}
            \centering
            \includegraphics[scale=.17]{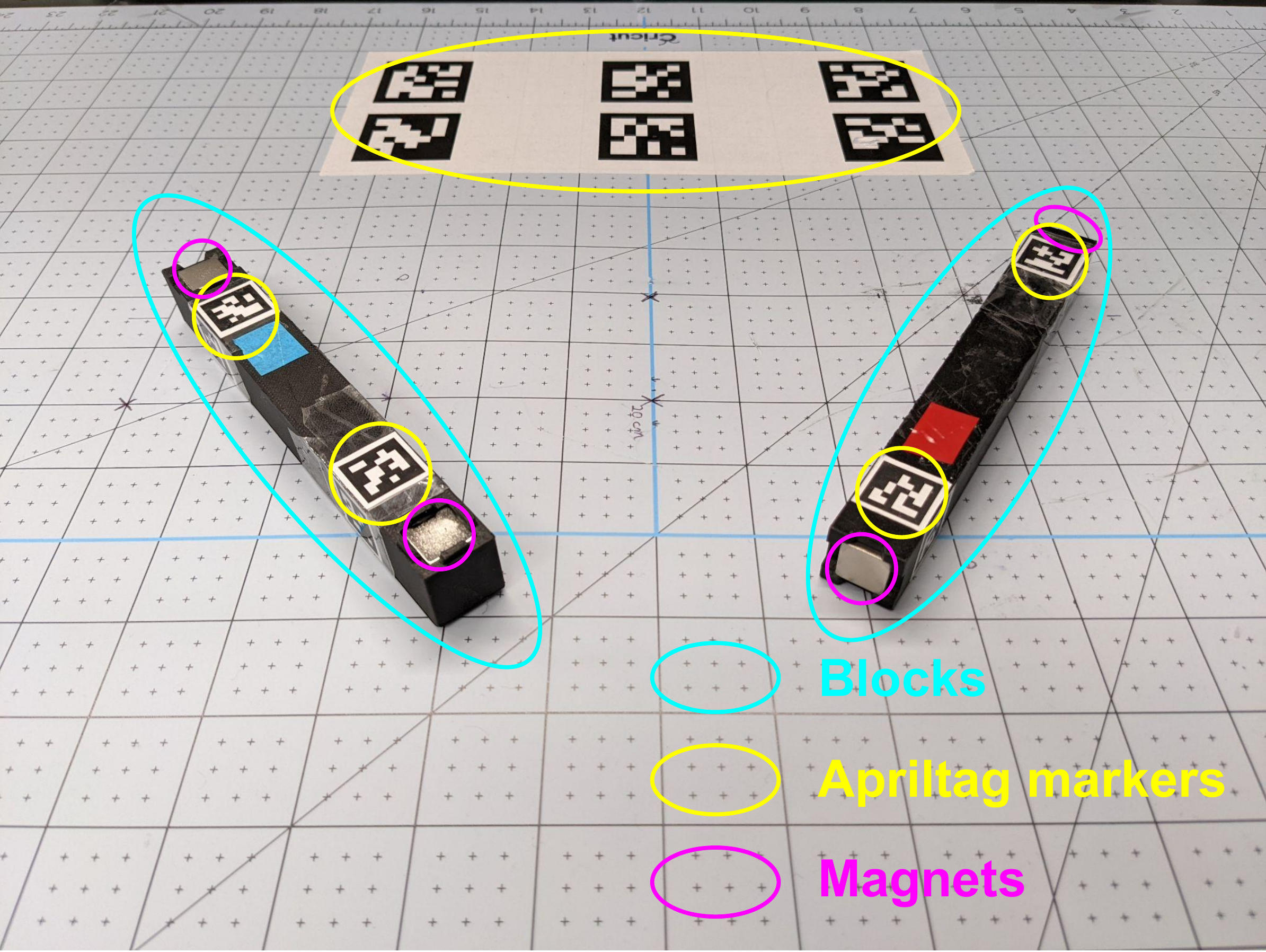}
            \caption{
                    Magnetic blocks and their components.
            }
            \label{fig:blocks}
        \end{minipage}\hfill
        \begin{minipage}{0.22\textwidth}
            \centering
            \includegraphics[scale=.17]{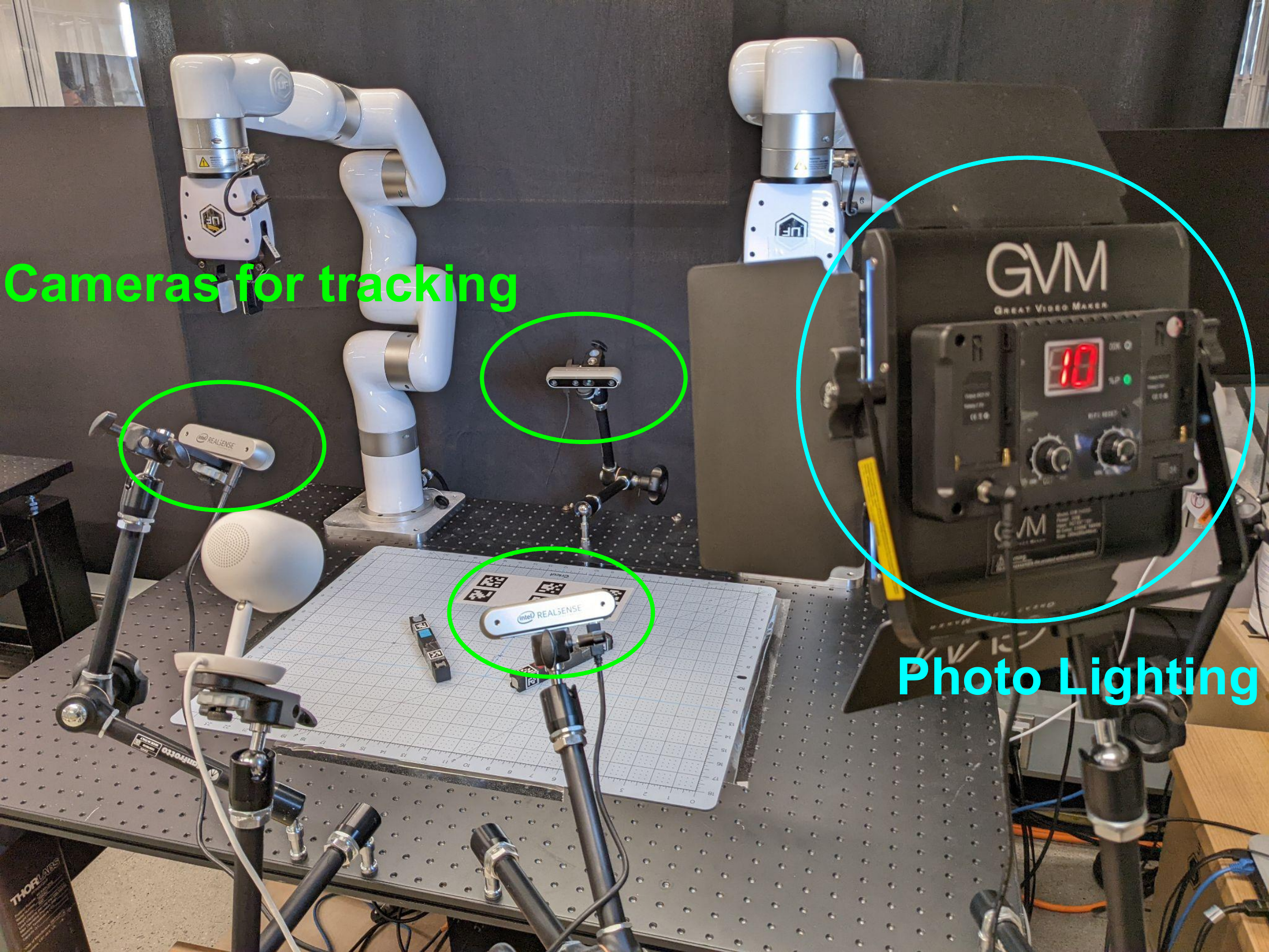}
            \caption{
                    Camera and lighting configuration.
            }
            \label{fig:cameras}
        \end{minipage}
        \vspace{-6mm}
    \end{figure}

    Both blocks also have the same arrangement of 8 Apriltag markers (15mm x 15mm). The blocks are printed by a 3D printer with PLA, each weighing 37.4g when including the two magnets. Figure \ref{fig:blocks} demonstrates the two blocks used in our work.
    
    \subsection{Peripherals}
    In addition to the robots and blocks we use the following peripherals.
        \paragraph{Tracking cameras} For tracking, we use three Intel RealSense D455 \cite{d455}. While the camera can provide depth images as well as RGB images, we only make use of the RGB images at a resolution of 1280 x 800 at 30fps. The produced images are used for Apriltag tracking system \cite{apriltag} and visualization of experiments. The cameras' position and directions can be arbitrarily configured as long as the Apriltag markers on the cutting mat are visible. Figure \ref{fig:cameras} demonstrates our camera placements.
        \paragraph{Recording cameras} For 24h/7 experiment recording, we use two Google Nest Indoor cameras. One nest camera is placed in front of the workspace, and the other in the right side of the workspace. Note that these cameras are not used for tracking.
        \paragraph{Photography lighting} To reduce variance of environmental lighting, we use a photography lighting kit to brighten the workspace in order to stabilize tracking. Although we found that we could still have successful estimation without this peripheral, the lighting diminishes the effect of changes in the ambient light.
        \paragraph{Workstation} We use a Lenovo P920 (CPU: xenon 6154 3.0GHz, Physical Memory: 188GB) to run our software stack including robot control and tracking.
    

\section{Software System for Physical Platform}
     \begin{figure}[t]
        \centering
        \includegraphics[width=\linewidth]{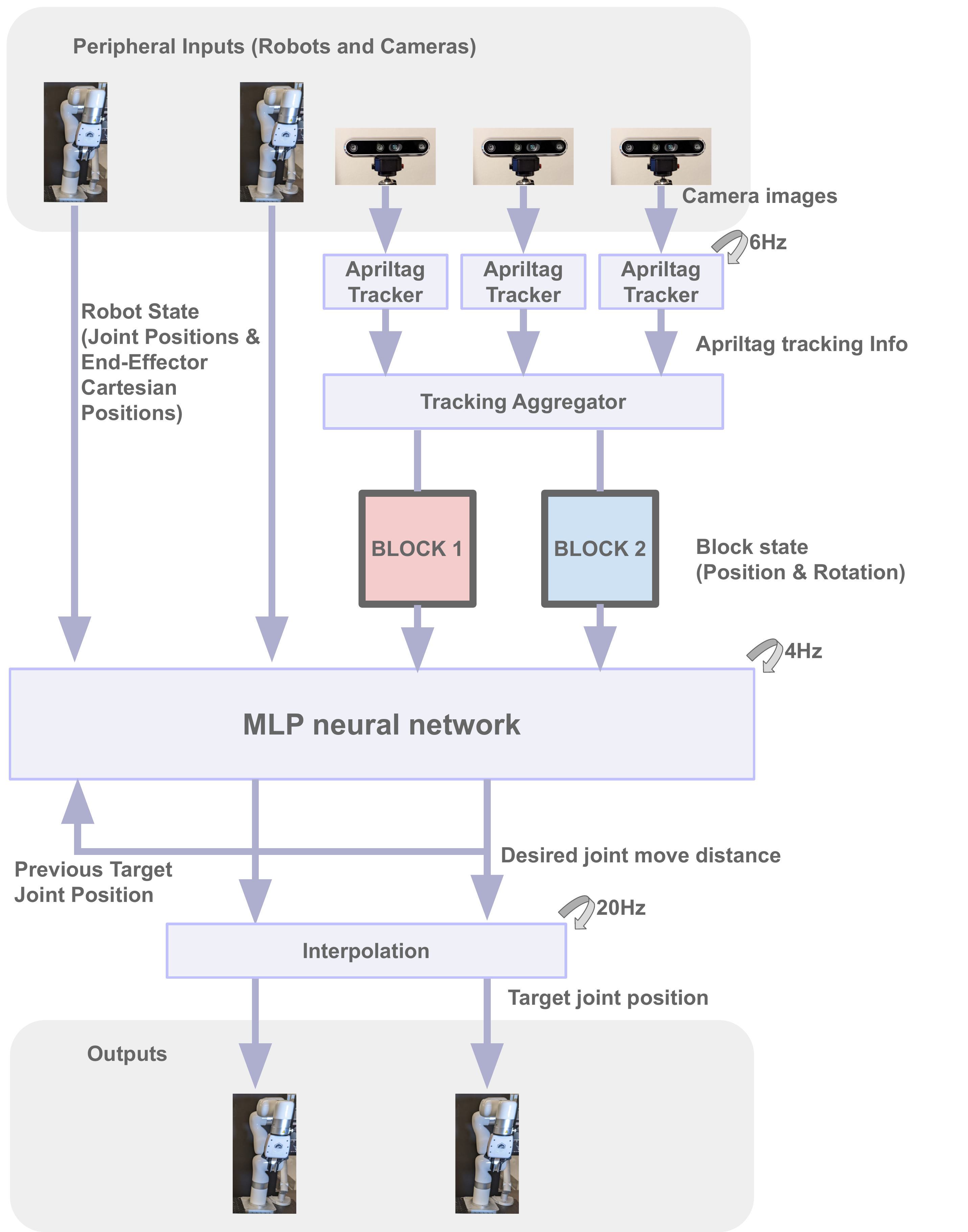}
        \caption{
            \small{
                Overall robot system architecture diagram.
            }
        }
        \vspace{-6mm}
        \label{fig:robot_system}
    \end{figure}
The software system for our physical platform consists of the following components 1) Object tracking and state-estimation 2) Neural network inference 3) Control interpolation, and is visually depicted in Figure \ref{fig:robot_system}.

\subsection{Object Tracking and State-Estimation}
In bi-manual manipulation tasks, robotics arms cause occlusions more frequently than a single arm setup, as the camera views can be covered by the two robotics arms. To address this issue, we use three cameras and run three tracking threads in parallel. Each tracking thread calculates the relative positions and orientations between base center and blocks by computing Perspective-n-Point (PnP) for detected Apriltag markers by using OpenCV library\cite{opencv}. 
These block states are aggregated and continuously updated by the latest states from the three threads.

\subsection{Neural Network Inference}
Since our system is designed to minimize the gap between simulation and real world, tracking estimates for the blocks, joint positions of two robotics arms, and the cartesian positions of the robotic end-effectors are directly fed into the neural network model. To improve the noise tolerance, we stack historical states for consecutive time steps as the input to the neural network model as explained in \ref{par:frame_stacking}. Specifically, the input for the neural network consists of: 1) position and 3x3 rotation matrices for two blocks, 2) 6 joint positions for each arms, and 3) positions of the robot end-effectors. The output of the neural network policy consists of: 1) displacements for each joint angle, and 2) binary signal on whether each gripper should open or close. As we run the inference step at 4Hz, the joint displacements are expected to be achieved in 0.25 seconds. 

\subsection{Interpolation}
To smoothly connect actions obtained from the agent at 4Hz, we use a cubic-spline interpolation to interpolate joint position targets from the inference component to generate joint commands at 20Hz.  We use same the interpolation in the training environment.


\subsection{Miscellaneous Utilities}
In addition to the three major software components (Tracking, Inference, and Interpolation), we have developed utilities that significantly improve our debugging and research iteration abilities. During real robot trials, we use camera viewers augmented with real-time object tracking overlays, while also visualizing the estimated state in our simulation environment. Additionally, we make use of continuous logging and low-level safety-checking running with a simulator to prevent robots and peripherals from being broken by unexpected behaviors by updating joint commands not to make collisions between robots, objects and specified boundaries defined in the simulator.

\section{Simulation Environment for Training Reinforcement Learning Policies}
\label{sec:sim}
In recent years, deep reinforcement learning (RL) has lead to many successes in a variety of domains \cite{go2016,dota2019,tokomak2022}, and in particular in the field of robotics \cite{qtopt2018, rubik2019}. However, training neural network policies through RL requires millions of environment interactions for even the simplest of tasks \cite{dmcontrol2018} \footnote{Indeed, as we will describe below, training policies for our ``Connect Task" requires 2 billion environment steps, which amounts to 5787 days of robot operation time.}. This, in addition to concerns of hardware cost and safety, makes RL policies infeasible to train directly in real-world setups. Thus, in this work we focus on how to train bimanual manipulation policies in a simulated setup, such that they can be successfully executed on real-world robotic systems.

Unfortunately, policies trained in simulation rarely transfer to real-world systems due to many factors such as model discrepancies and noise in real-world state-estimation. This problem -- often referred to as the Sim2Real problem -- has lead to a significant body of work in recent literature \cite{sim2real2017,sim2real2018,rubik2019}. In this section, we outline how we have designed our simualted environments to enable successful transfer of RL policies from simulation to real-world robots.




\subsection{Direct Joint-Space Control}
    In this work we have decided to train policies for real-time (4Hz) joint velocity control of the robot arms. As opposed to catesian control of end-effector position, joint-space control not only alleviates the need for inverse kinematics (IK),
    but enables policies to fully utilize the joint space to learn bimanual behaviors that avoid collisions between the robot arms while executing the desired task.

\subsection{Behavior Constraints}
    \paragraph{Joint Velocity and Acceleration Limit}
    To enable joint space policies to produce feasible joint position targets in the real world, it is important to reproduce strict constraints for joint velocity and acceleration limits in simulation. To this end, the policy outputs -- which are target joint velocities for the robotic arms -- are clipped in order to satisfy the joint acceleration limits, details of which are presented in Appendix \ref{app:acc_constraint}.
    
    \paragraph{Applied Force Limit}
    Excessive applied forces to the robots can cause robots to fault in the real world. 
    Additionally, high applied forces may easily lead to object breakages in real world, even if the applied forces are within the robots' configuration limits. To encourage policies to learn less aggressive behaviors and avoid excessive forces,
    in each simulation step we calculate applied forces for all contacts and add negative rewards for violations of force limits. This limitation helps the robots avoid pushing gripped objects to the ground, which can lead to robots' faulting due to joint load limits.
    In the Mujoco \cite{mujoco2012} simulator, information regarding applied forces can be obtained through a ``constraint force" function.

\subsection{Perception and Actuation Noise}
\begin{figure}[t]
            \centering
            \includegraphics[scale=.5]{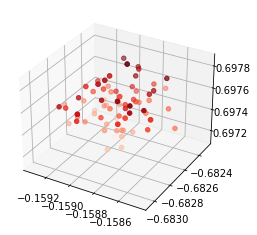}
            \caption{
                \small{
                    Observations of real-world 3D block position used to calibrate observation noise level.
                }
            }
            \vspace{-6mm}
            \label{fig:perception_noise}
        \end{figure}

    \paragraph{State Estimation Noise}
    In the real world, perception data from sensors are inevitably noisy. To address this problem, instead of using statistical filters (e.g. Kalman filter) at runtime, we opted for incorporating noise into our simulation. Specifically, we applied a zero-mean spherical Gaussian noise to both the positions and the orientations of the two blocks. Based on data collected in the real world (Figure \ref{fig:perception_noise}), we used 10mm as the standard deviation of the position noises, and 0.025 rad as the standard deviation of the rotational noises.
    Incorporating noise directly into our simulation environment prevents the policy from learning highly agile behaviors that take advantage of idiosyncracies of simulator physics, and results in more conservative policy behaviors that are robust to observation noise.
    Additionally, this allows us to remove an extra abstraction layer for filtering sensor data in real world,
    which not only contributes to the simplicity of the system, but also reduces the gap between the simulation and real world software stacks.
    
    \paragraph{Joint Acceleration Noise}
    In the real-world, there exists a discrepancy between commanded and output accelerations of robot joints due to a number of factors such as current robot pose and electric current output. As can be seen in Figure \ref{fig:pick_acc} which shows the output acceleration is rapidly oscillating in real, we find that in addition to acceleration limits, applying Gaussian noise to commanded accelerations in simulation (details in Appendix \ref{app:acc_constraint}) results in policies that learn to have more faviorable acceleration behavior.

\section{Agents}
    \label{sec:agents}
    In this section we describe our agents, including observations and action spaces, and MLP policy network architecture.
    \subsection{Policy Observations and Actions}
    The observations provided to the agent can be divided into two broad categories, those concerning the blocks, and those concerning the robots. When designing observations to provide to agents, we have taken the effort to ensure the same observations are available in both simulation and the real world.
    \paragraph{Block Observations}
    We use very minimal block observations, namely the global position and orientation (represented as a 3x3 rotation matrix) for all blocks. In the real world environment, we simply transform tracking results from our Apriltag tracking system for the blocks to match our simlator's coordinate system, without using any additional statistical estimation filtering, and directly feed them into the policy.
    \paragraph{Robot Observations}
    For each robotic arm we include, current joint positions, current end-effectors position, target joint positions in the previous timestep, and the extent of gripper opening in the previous timestep. In the real world environment, these states are obtained from the robot controllers. The end-effector positions are transformed to match simulator's coordinate system before being fed into the policy.
    \paragraph{Frame Stacking}
    \label{par:frame_stacking}
    To improve tolerance to noises incorporate into the simulator, we provide history of the past by including observations from 8 previous environment steps.
    \paragraph{Robot Action}
    Our agents' action output consists of joint velocities and gripper opening positions for two robotics arms.
    
    \subsection{Policy and Critic Networks}
    Observations are fed to a neural network policy which consists of a 4 layer MLP, with 1024 dimensional hidden layers and swish non-linearities. The output of the policy network consists of two vectors representing the mean and standard deviations for a diagonal multivariate Gaussian distribution.
    A sample from this distribution is then squashed by an elementwise Tanh function, and used as the control action.
    To train our RL agents, we also require critic value estimates, which we obtain by passing observations to a critic neural network. The critic network uses an identical architecture as the policy network, with the difference that the critic network's output is a real-valued scalar representing the estimated value.

\section{Training and Evaluation}
\subsection{Large-Scale PPO}
    Our RL policies are trained using Proximal Policy Optimization (PPO) \cite{schulman2017proximal} and Generalized Advantage Estimation (GAE) ~\cite{schulman2015high}, and follow the practical PPO training advice of \cite{andrychowicz2020matters}. As will be shown below, a key ingredient in enabling the training of our bi-manual object manipulation agents is the scale of training. Our agents are trained for 2-3 billion environment steps, using 1 Nvidia V100 GPU for training, and 3000 preemptible CPUs for generating rollouts in the environment. 1 Billion steps in our setup amounts to about 10 hours of training. The key libraries used for training are Jax~\cite{jax2018github}, Haiku~\cite{haiku2020github}, and Acme~\cite{hoffman2020acme}. We use Mujoco\cite{mujoco2012} as the simulator for the training environment. In each training episode, the simulated environment starts with the blocks randomly dispersed on the ground, and robots are reset to predefined initial poses as shown in Figure \ref{fig:connect_task}.
\subsection{Training Performance Evaluation}
    During the training, we evaluate trained policies continuously, approximately every 10 minutes, by freezing the policy and computing 
    average success rate over 40 episodes. This continuous evaluation is executed on a dedicated CPU in parallel to the training processes. Additionally, in each evaluation cycle we generate a video to visualize the agents' current learned behavior. This generates a better understanding of training performance and has been an invaluable tool in debugging and research iteration.

\section{Experiments}
\subsection{Pickup Task Experiment}
    \label{sec:pick}
    \begin{figure*}[t]
        \begin{minipage}{0.5\textwidth}
            \centering
            \includegraphics[scale=.17]{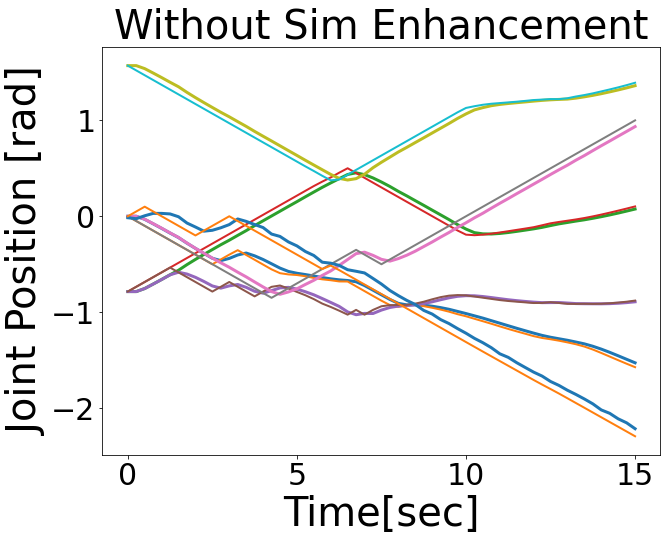}
            \includegraphics[scale=.17]{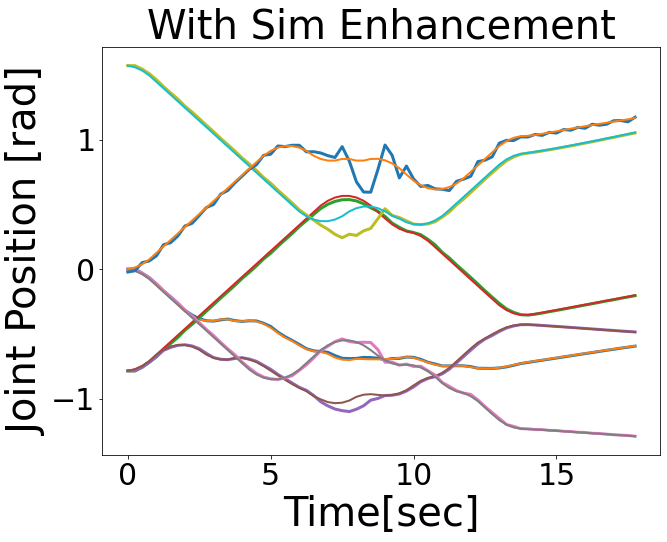}
            \caption{
                \small{
                    Pickup Task: All Generated Joint Position Trajectories
                }
            }
            \label{fig:pick_all_traj}
        \end{minipage}
        \hfill
        \begin{minipage}{0.5\textwidth}
            \centering
            \includegraphics[scale=.165]{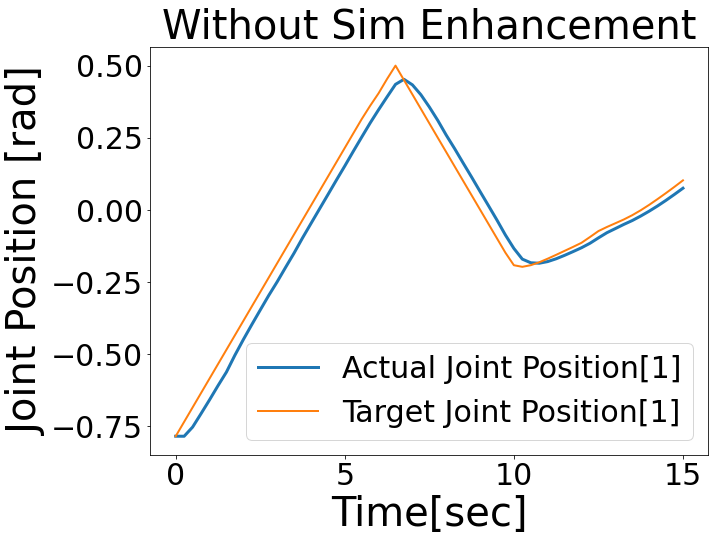}
            \includegraphics[scale=.165]{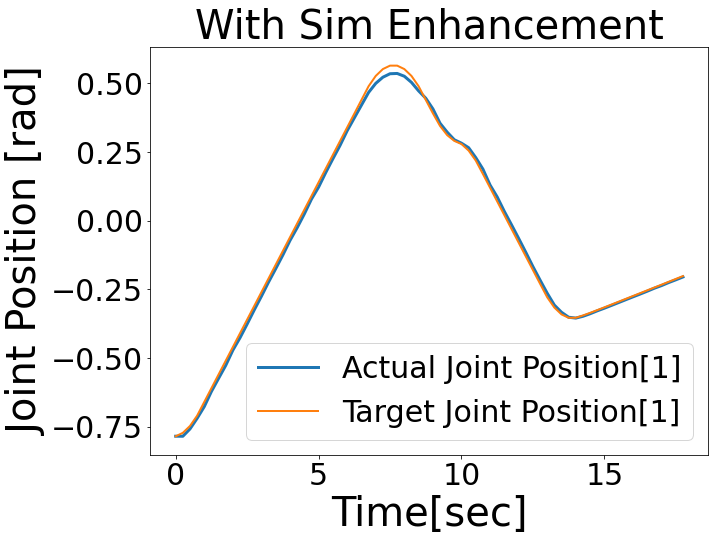}
            \caption{
                \small{
                    Pickup Task: Joint Position Comparison
                }
            }
            \label{fig:pick_one_traj}
        \end{minipage}
        
        \vspace{4mm}
        
        \begin{minipage}{0.5\textwidth}
            \centering
            \includegraphics[scale=.17]{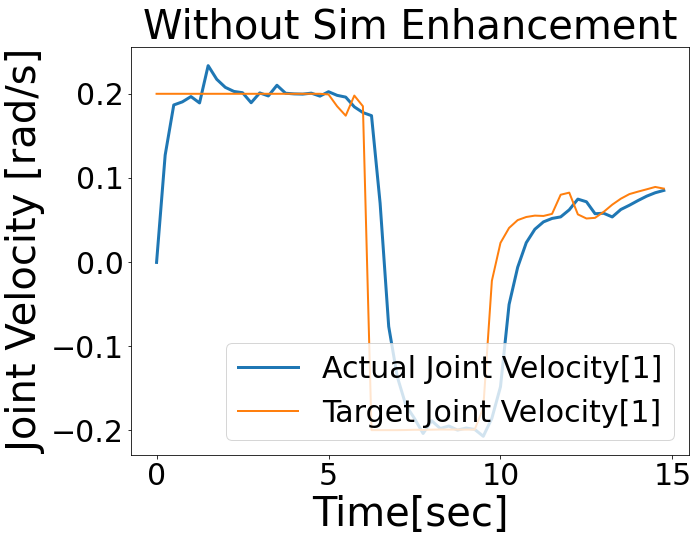}
            \includegraphics[scale=.17]{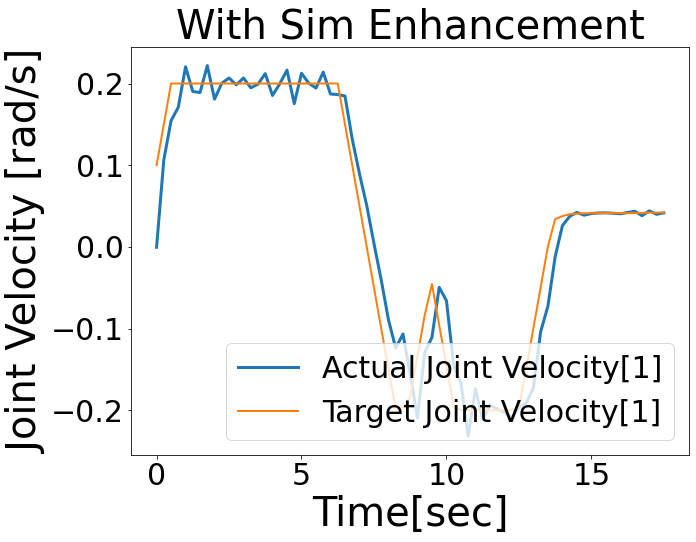}
            \caption{
                \small{
                    Pickup Task: Joint Velocity Comparison
                }
            }
            \label{fig:pick_vel}
        \end{minipage}
        \hfill
        \begin{minipage}{0.5\textwidth}
            \centering
            \includegraphics[scale=.17]{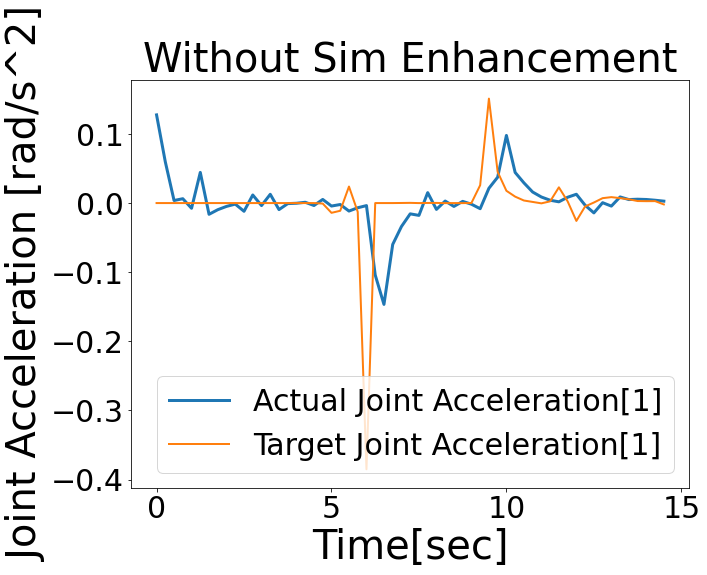}
            \includegraphics[scale=.17]{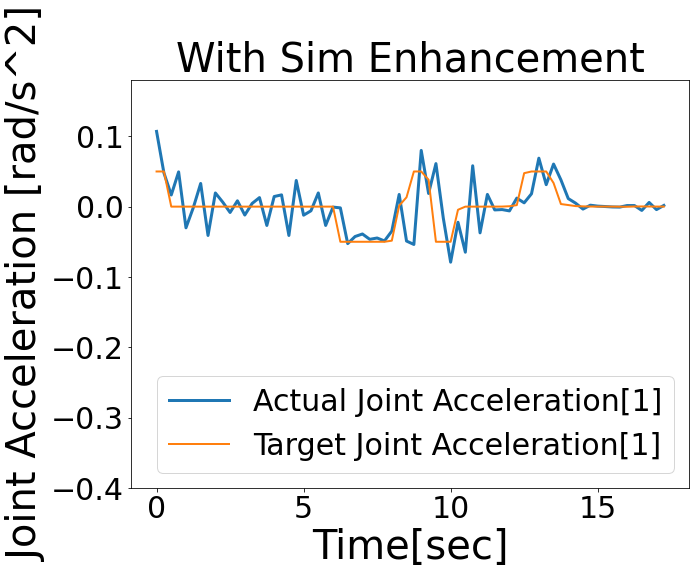}
            \caption{
                \small{
                    Pickup Task: Joint Acceleration Comparison
                }
            }
            \label{fig:pick_acc}
        \end{minipage}
        
        \begin{minipage}{0.5\textwidth}
            \centering
            \begin{table}[H]
                \scriptsize
                \centering
                \begin{tabular}{|c|c|c|c|c|}
                    \hline
                    \multicolumn{1}{|c||}{Training Environment} & \multicolumn{2}{c|}{Success $(<5mm)$} & \multicolumn{2}{c|}{Failure}  \\ \cline{4-5}
                    \multicolumn{1}{|c||}{} & \multicolumn{2}{c|}{} & Gripped & No-Grip \\ \hline
                    \multicolumn{1}{|c||}{Without enhancements} & \multicolumn{2}{c|}{0.05} & \multicolumn{2}{c|}{0.95} \\ \cline{4-5}
                    \multicolumn{1}{|c||}{} & \multicolumn{2}{c|}{} & 0.75 & 0.2 \\ \hline
                    \multicolumn{1}{|c||}{With enhancements} & \multicolumn{2}{c|}{1.00} & \multicolumn{2}{c|}{0.0} \\ \hline
                \end{tabular}
                \caption{
                    \small{
                        Success Rate for Pickup Task in Real World:
                    }
                    \footnotesize {
                        Success rates are calculated after 20 consecutive trials.
                        We also report whether failure cases were due to inability to grip, or inability to move the block to desired position.
                    }
                }
                \label{tab:pick}
            \end{table}
        \end{minipage}
        \hfill
        \begin{minipage}{0.5\textwidth}
            \centering
            \begin{table}[H]
                \scriptsize
                \centering
                \begin{tabular}{|c|c|c|c|c|c|}
                    \hline
                    \multicolumn{1}{|c||}{Training Environment} & \multicolumn{3}{c|}{Success} & \multicolumn{2}{c|}{Failure}  \\ \cline{2-4} \cline{5-6}
                    \multicolumn{1}{|c||}{} & 1mm & 5mm & 10mm & Gripped & No-Grip \\ \hline
                    \multicolumn{1}{|c||}{Without enhancement} & \multicolumn{3}{c|}{0} & \multicolumn{2}{c|}{1.0} \\ \cline{2-4} \cline{5-6}
                    \multicolumn{1}{|c||}{} & 0 & 0 & 0 & 0.6 & 0.4 \\ \hline
                    \multicolumn{1}{|c||}{With enhancement} & \multicolumn{3}{c|}{0.65} &\multicolumn{2}{c|}{0.35} \\ \cline{2-4} \cline{5-6}
                    \multicolumn{1}{|c||}{} & 0.35 & 0.15 & 0.15 & 0.35 & 0 \\ \hline
                \end{tabular}
                \caption{
                \small{Success Rate for Connect Task In Real World:}
                                \footnotesize {
                    Success rates are calculated after 20 consecutive trials. We set 1mm/5mm/10mm as different levels of success.
                    We also report whether failure cases were due to inability to grip, or inability to move the block to desired position.
                }
                }
                \label{tab:connect}
            \end{table}
        \end{minipage}
        \vspace{-4mm}
    \end{figure*}

    To verify feasibility of real-time joint space control using RL, we begin with the ``Pickup Task" as described in Section\ref{sec:pickup_task}.
    As we note in Section \ref{sec:large_scale_training}, agents can learn the ``Pickup Task" about 10 times faster than the ``Connect Task". Therefore, the ``Pickup Task" is useful to analyze basic joint behaviors and simulation enhancement parameters before moving onto the ``Connect Task". In Section \ref{sec:sim} we described a number of modifications that we made to our simulated environments to enable the transfer of learned policies to our real-world setup. Here, we compare training our agent with and without our proposed simulation enhancements.
    
    For the 6 robot joints, Figure \ref{fig:pick_all_traj} compares the overlap of desired joint position trajectories generated by our agent, and the actual joint position trajectories observed during execution, with and without simulator enhancements. For closer inspection, Figure \ref{fig:pick_one_traj} shows compares the trajectory for the first joint. As can be seen, the agent without simulator enhancements generates more aggressive trajectories than the agent trained with simulator modifications. Since the real robot is unable to produce the aggressive behaviors, there is a gap between the target joint position and the actual joint position when trained without enhancements. We also present the difference of velocity trajectories in Figure \ref{fig:pick_vel}, and the difference of acceleration trajectories in Figure \ref{fig:pick_acc}. These Figures indicate that the agent generates less aggressive behaviors when trainin RL policies using our modified simulator.
    
    We now study the effect of simulation enhancements on agents' success rate in the ``Pickup Task".
    As can be seen in Table \ref{tab:pick}, we find that while agents trained without enhancement can successfully grip the block 75\% of the time, they cannot move the block to the specified position for in most cases (95\% failure rate). When simulator enhancements are incorporated, the agent succeeds at the ``Pick Up" task with 100\% success rate
    Interestingly, we observed the agent can retry picking up the block even after the block has dropped from the gripper which means our agent can generate robust behavior even in accidental situations.
    These results show that the simulation enhancement improves both 1) trajectory tracking performance, and 2) positioning accuracy under environmental noises in the real world.
    


\subsection{Connect Task Experiment}
    \begin{figure*}[t]
        \begin{minipage}{\textwidth}
            \centering
            \includegraphics[scale=0.5]{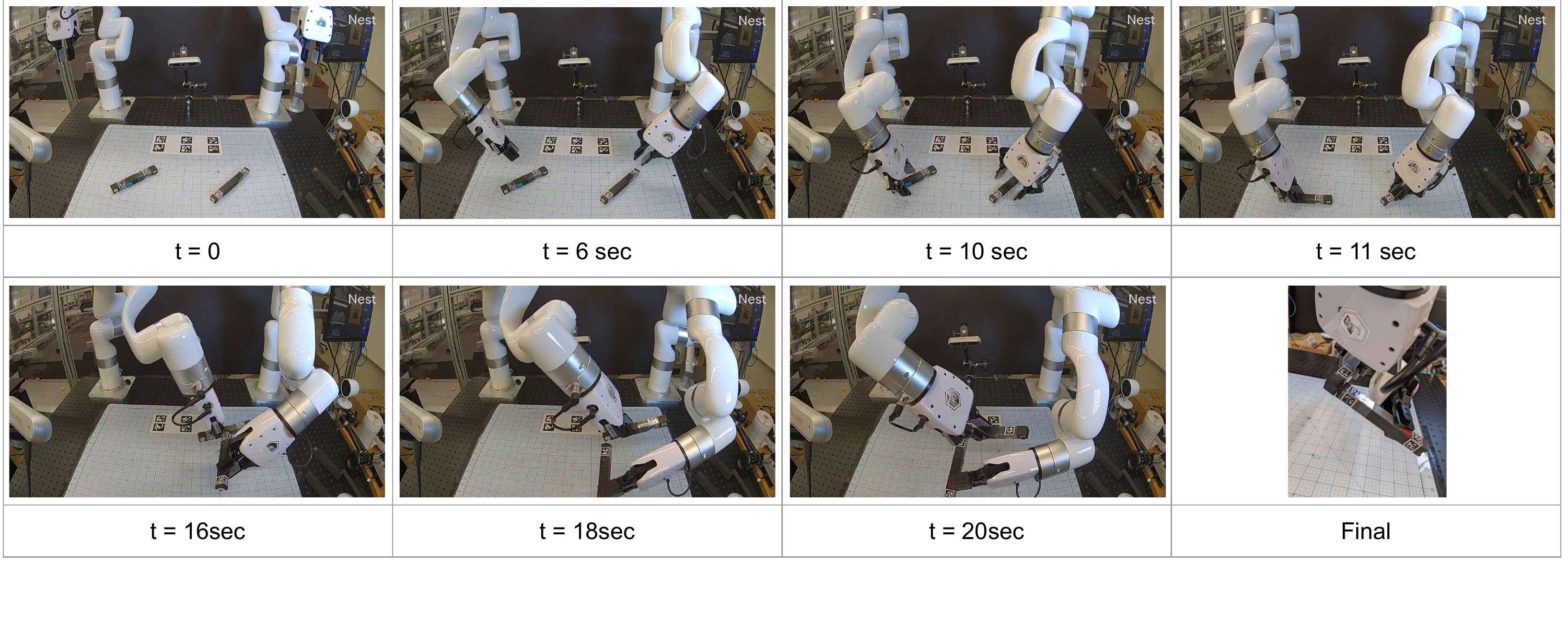}
            \vspace{-8mm}
            \caption{
                \small{
                    Snapshots For Connect Task Execution in Real World.
                }
            }
            \label{fig:connect_photos}
            \vspace{4mm}
        \end{minipage}
        \begin{minipage}{0.5\textwidth}
            \centering
            \includegraphics[scale=.17]{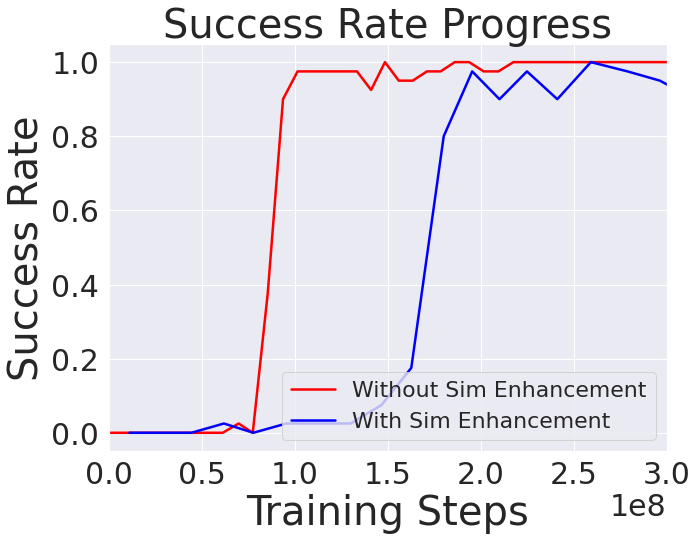}
            \includegraphics[scale=.17]{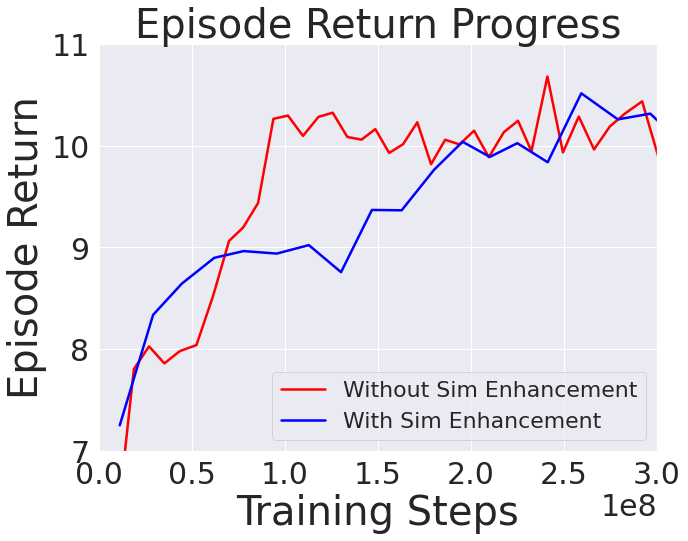}
            \caption{
                \small{
                    Pickup Task Training Progress.
                }
            }
            \label{fig:pick_progress}
        \end{minipage}
        \hfill
        \begin{minipage}{0.5\textwidth}
            \centering
            \includegraphics[scale=.17]{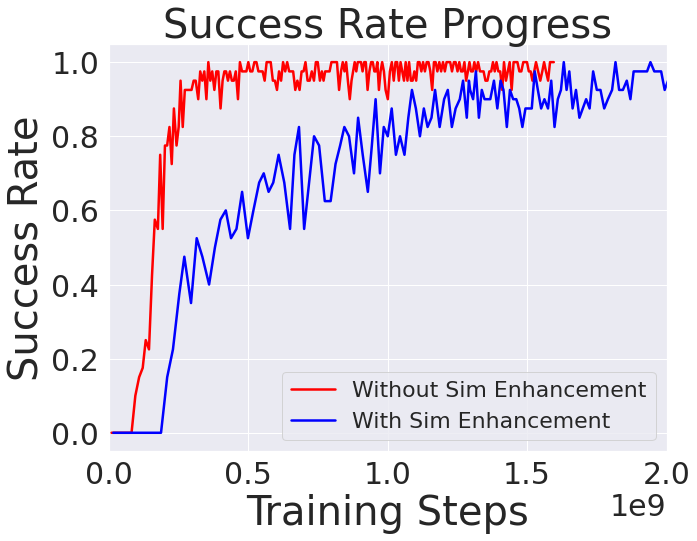}
            \includegraphics[scale=.17]{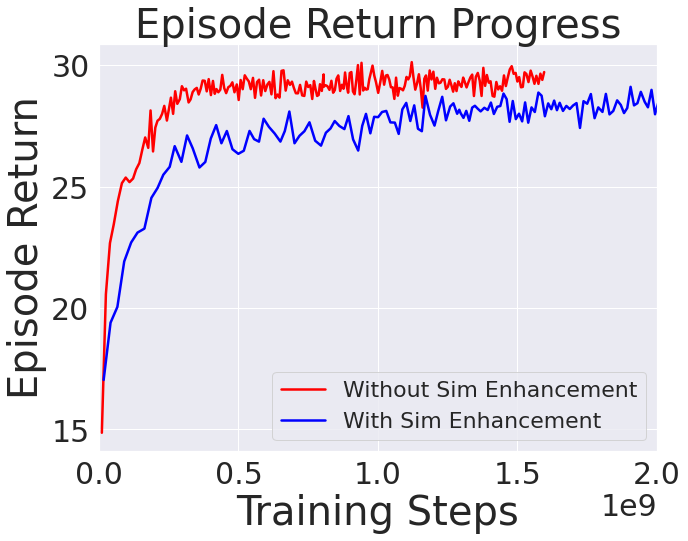}
            \caption{
                \small{
                    Connect Task Training Progress.
                }
            }
            \label{fig:connect_progress}
        \end{minipage}
    \end{figure*}
    
\begin{figure}[t]
        \centering
        \includegraphics[scale=.17]{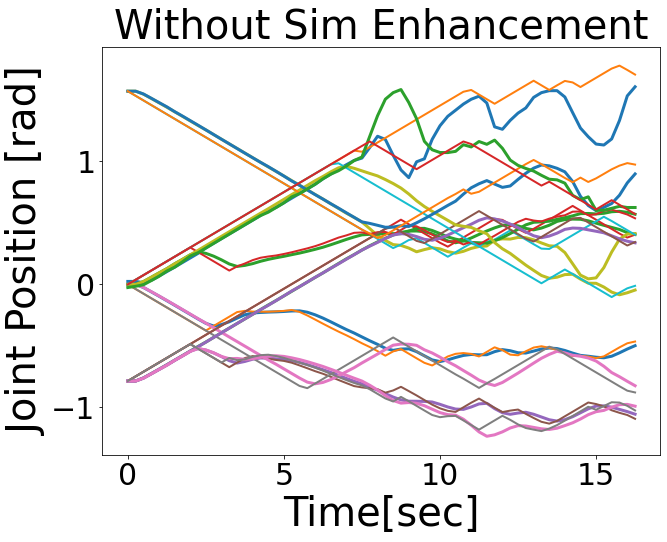}
        \includegraphics[scale=.17]{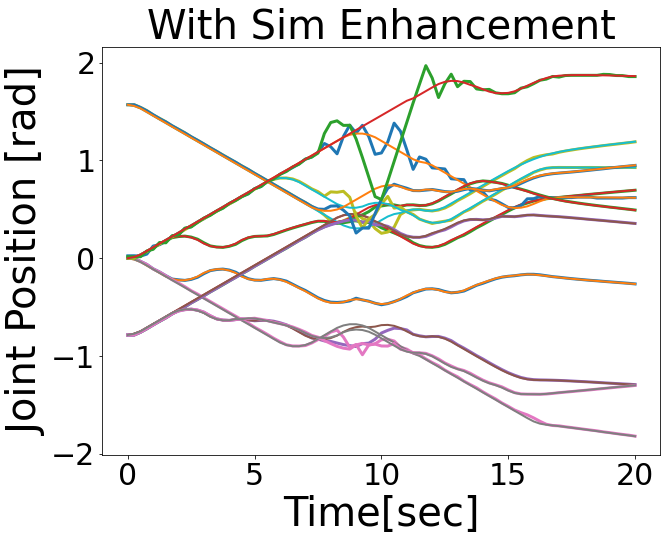}
        \caption{
            \small{
                Connect Task: All Generated Joint Position Trajectories
            }
        }
        \label{fig:connect_all_traj}
    \end{figure}

    To evaluate the contribution of our simulation enhancement for bi-manual collaborative object manipulation, we compare the connect task's success rate of two agents, one trained with and one trained without our simulation enhancement in Table \ref{tab:connect}. Without enhancement, the agent can grip blocks at 60\% but can't move them to the connected position. In contrast, the agent with enhancement can grip blocks at 100\% and can connect the blocks at 35\%, can move the blocks to the connecting position within 5mm gap at 15\% and can move the blocks to the connection position within 10mm gap at 15\%. Figure \ref{fig:connect_photos} shows snapshots of the behaviors in the real world execution, generated by the agent with enhancement, which are corresponding to the trajectories in Figure \ref{fig:connect_all_traj}. In the snapshots, the two robotics arms accurately grip the blocks and move the blocks to the connecting position and keep connected for three seconds. This result shows that the agent with enhancement can acquire the ability to generate behaviors to keep the successful goal state with real world perception noises. While we observed lower rate successes of pickup task by the agent without enhancement, the agent can't make any successful demonstration without enhancement. This result indicates our enhancement gives the agent an ability to learn higher complexity tasks for the real world execution. This is an exciting finding as our agent successfully demonstrates joint control for bi-manual collaborative object manipulation.
    
\subsection{Importance of Large-Scale Training}
    \label{sec:large_scale_training}
    We train our MLP agent (Section \ref{sec:agents}), for 3.2 billion environment steps to observe training patterns that may arise over a long period of training. In Figures \ref{fig:pick_progress} and \ref{fig:connect_progress} we compare training RL policies with and without simulator modifications that were incorporated towards making learned policies transferable to real-world robots (e.g. actuator constraints and observation noise as described in Section \ref{sec:sim}). As can be seen, for both tasks, simulator modifications make training require significantly more iterations to achieve a high success rate.
    For the ``Pickup Task", success rate exceeds 90\% at 100 million steps without simulator modifications, while taking 190 million steps when they are incorporated.
    In the ``Connect Task", success rate exceeds 90\% at 280 million steps without simulation modifications, while taking 1.8 billion steps when they are incorporated. Although agents train faster without simulation modifications, as discussed in the above sections, they learn less robust policies that do not transfer well to real-world robots.
\section{Related Works}
\textbf{Construction:}
\cite{bapst2019structured} previously studied two-dimensional construction environments, training an agent to assemble a structure for an open-ended goal, such as a connecting or a covering structure. Certain details of assembly are abstracted away, as the agent has the ability to directly summon a block of choice anywhere in the scene and weld blocks via an explicit action. \cite{lee2019ikea,lee2021adversarial} also introduce a three-dimensional assembly environment for furniture design from a blueprint. By contrast, our environment contains all of the necessary features for the real world execution (e.g. physical blocks, precise robotics model, noises and actuation constraints). Also, we successfully executed real-time joint space control by our agent in the real world to construct a structure where two magnets are connected.
\cite{suarez2018can} studied assembly of a single chair with real-world bimanual robots using offline planning methods. By contrast, we use a neural network policy which generates online actions. As mentioned in Section \ref{sec:pick}, the agent can produce robust behavior which retries picking after the robot accidentally drop the blocks.

\textbf{Generalization in Robot Manipulation:}
Much recent work in robot manipulation focused on the tasks of object grasping \cite{levine2018learning,kalashnikov2018qt}, in-hand object manipulation \cite{andrychowicz2020learning,chen2021system,huang2021generalization}, or execution of a motor skill \cite{yu2020meta}, where variation comes from diversity of object shapes and arrangements involved. By contrast, we use two robotics arms and generates joint trajectories directly from the agent which can handle noisy perception and actuation through the neural network in real world.

A key point of differentiation from the above works is that we use bi-manual object manipulation as a domain for generalization for sim to real transfer, and train a neural network policy for semi real-time joint control that can solve the tasks with noisy perception and actuation in the real world without needing additional layers for collision checking, perception filtering, inverse kinematics and trajectory generation apart from behaviors produced by the simulator.

\section{Conclusion}

In this work, we proposed the connection of magnetic blocks as a minimal task for studying bimanual object manipulation. Despite its simplicity, the ``Connect Task" stresses key components of bimanual object manipulation such as coordination for task execution as well as collision avoidance between the two robot arms. Additionally, our proposed setup can be readily extended to more complex bimanual manipulation tasks, such as building desired structures from a collection of magnetic blocks.

Given the complexity of obtaining effective bimanual robot controllers, we then studied how to obtain effective policies through simulator-trained reinforcement learning such that they successfully transferred to a real-world robotic setup. We discussed in detail how our RL approach, with simple modifications to our simulator, significantly simplifies the process of obtaining robot policies. Specifically, we discussed how the use of joint space control removes the need for inverse kinematics and enables neural networks to learn collision-free robot behaviors, and how simulator modifications handle actuation noise and obviate the need for observation filtering. We also highlight how the above choices not only simplify on-robot software layers, but also reduce the gap between simulation and real-world environments, which improves training accuracy and transfer to real-world systems.

\section{Appendix}
\subsection{Acceleration Constraint and Noise Calculation}
\label{app:acc_constraint}
\begin{lstlisting}[language=Python, basicstyle=\tiny]
def filter_velocity(target_position_diff: np.ndarray,
                    prev_position_diff: np.ndarray,
                    dt: float,
                    acceleration_limit: float,
                    noise_scale: Optional[float] = None) -> np.ndarray:
  noise_scale = noise_scale or 0.0
  target_velocity = target_position_diff / dt
  current_velocity = prev_position_diff / dt
  # logging.info(f"tvel = {target_velocity}, avel = {current_velocity}")
  desired_acc = target_velocity - current_velocity
  clipped_acc = np.clip(desired_acc, -acceleration_limit * dt,
                        +acceleration_limit * dt)
  clipped_acc += noise_scale * clipped_acc * np.random.normal(clipped_acc.shape)
  # logging.info(f"aacc with noise = {clipped_acc}")
  next_velocity = current_velocity + clipped_acc
  return next_velocity * dt

\end{lstlisting}


\end{document}